# Thai Semantic End-of-Turn Detection for Real-Time Voice Agents

Thanapol Popit*
*Department of Computer Engineering*
KMUTT
Bangkok, Thailand
thanapol.popi@kmutt.ac.th

Natthapath Rungseesiripak
*Innovation Lab*
SCBX
Bangkok, Thailand
natthapath.r@scbx.com

Monthol Charattrakool
*Innovation Lab*
SCBX
Bangkok, Thailand
monthol.c@scbx.com

Saksorn Ruangtanusak
*R&D*
SCBX
Bangkok, Thailand
saksorn.r@scbx.com

*Abstract*—Fluid voice-to-voice interaction requires reliable and low-latency detection of when a user has finished speaking. Traditional audio silence end-pointers add hundreds of milliseconds of delay and fail under hesitations or language-specific phenomena. We present, to our knowledge, the first systematic study of *Thai* text-only end-of-turn (EOT) detection for real-time agents. We compare zero-shot and few-shot prompting of compact LLMs to supervised fine-tuning of lightweight transformers. Using transcribed subtitles from the YODAS corpus and Thai-specific linguistic cues (e.g., sentence-final particles), we formulate EOT as a binary decision over token boundaries. We report a clear accuracy–latency trade-off and provide a public-ready implementation plan. This work establishes a Thai baseline and demonstrates that small, fine-tuned models can deliver near-instant EOT decisions suitable for on-device agents.

## I. INTRODUCTION

Semantic End-of-Turn Detection (EOT) is the task of predicting whether a speaker has finished their conversational turn using purely the linguistic content of their utterance. Unlike traditional methods that rely heavily on acoustic cues like silence duration (pauses), such as the widely used Silero Voice Activity Detector (VAD) [1], this approach analyzes the transcribed text to understand if the sentence or thought is semantically complete. The primary importance of this task is to reduce latency and improve the naturalness of human-computer interactions. By accurately predicting the end of a user's turn before they fall silent, a conversational agent can begin processing the request and formulating a response immediately, leading to faster, more fluid, and less awkward conversations [2]–[4].

While research in Thai NLP has explored tasks such as sentence segmentation in written text, the specific problem of predicting conversational turn completion in spoken dialogue remains largely unaddressed. To the best of current knowledge, no prior work has explicitly tackled semantic end-of-turn detection for Thai. This highlights a substantial research gap and presents an opportunity to extend dialogue modeling efforts to a linguistically and culturally distinct context.

*Work conducted as part of the SCBX Summer Internship program.

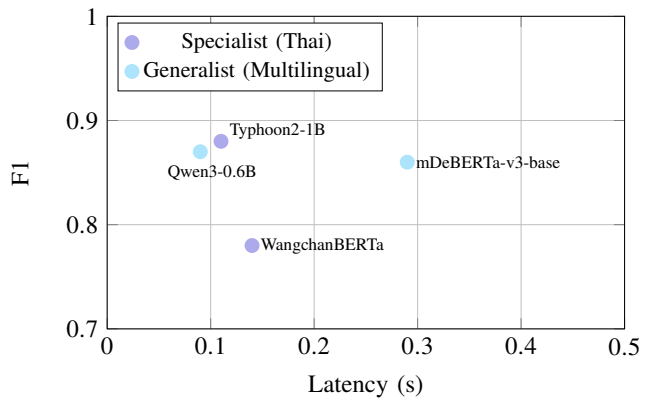

Fig. 1: Accuracy–latency trade-off for Thai EOT detection models (all Fine-tuned). Purple markers denote specialist (Thai-specific) models; blue markers denote generalist (multilingual) models.

We study *text-only* EOT: given a speech-to-text transcription, decide if the user intends to finish their conversational turn. Text-only EOT avoids tuning acoustic thresholds, reduces perceived delay, and aligns with sentence boundary detection. Our contributions:

- **First Thai EOT benchmark.** A formal task and dataset recipe for Thai EOT using public subtitles, with Thai-specific analysis.
- **Fine-tune vs zero-/few-shot comparison.** We assess compact LLMs in zero/few-shot mode against fine-tuned encoders/decoders under the same evaluation protocol, emphasizing CPU latency (see Fig. 1).

## II. RELATED WORK

### A. End-of-Turn Detection

Prior work often models acoustic-prosodic cues with auxiliary intent prediction [5]. Industrial EOT detectors typically gate on silence durations [1], [11]. Recent open-source systems learn semantic turn cues directly from audio (Smart Turn v2) or from conversation context (TEN Turn Detection) [8]–[10]. Turnsense fine-tunes a 135M-parameter transformer for End-of-Utterance (EOU) detection [7]. TurnGPT [13] demonstrated that Transformer-based LMs trained on dialogue corpora can predict likely turn shifts, outperforming RNNs. Similarly, Masumura et al. (2018) [19] proposed dual LSTM models for Japanese dialogues, showing context from both speakers improves EOT detection. Aldeneh et al. (2018) further showed that predicting speaker intentions as an auxiliary task boosts EOT performance [5]. These works confirm that contextual language modeling is beneficial for turn-taking tasks.

### B. Decoder-Only Based EOT Detection

Transformer architectures underpin modern LLMs and specialized EOT detectors. Within this paradigm, decoder-only models can be leveraged in two primary ways: zero-shot/few-shot prompting and fine-tuning. For zero-shot EOT detection, compact models like Qwen3-0.6B [23] and Llama3.2-Typhoon2-1B [24] serve as efficient generative backbones. While their larger, instruction-tuned counterparts (Qwen3-8B, Llama3.1-Typhoon2-8B) generally provide superior performance, they do so at the cost of greater latency.

Alternatively, fine-tuning these models for EOT marks a notable conceptual shift. Rather than framing the task as a discrete binary classification problem, this approach models turn completion as a continuous, probabilistic prediction. [13], [14] Such a formulation aligns naturally with the incremental, real-time dynamics of spoken dialogue. In this view, the system is not explicitly designed as an "EOT detector"; instead, a generative language model implicitly learns the statistical likelihood of a turn-shift at any point in the conversation.

The underlying principle is to treat the end of a conversational turn as simply another predictable event within a sequence of language. [12] This is operationalized by introducing a special token into the model's vocabulary. During fine-tuning on a dialogue corpus, this token is inserted at speaker-change boundaries. With a standard autoregressive next-token prediction objective, the model learns to assign high probabilities to the EOT token when a turn is semantically complete. [13] In a real-time interaction, the likelihood assigned to this special token at each step is interpreted as the probability that the speaker's turn has concluded.

This fine-tuning approach also presents a trade-off between performance and efficiency. Smaller variants, such as a fine-tuned Qwen3-0.6B, are ideal for on-device applications where low latency is critical. In contrast, mid-sized models like a fine-tuned Llama3.1-Typhoon2-8B can strike a more effective balance between predictive accuracy and resource consumption.

### C. Encoder-Only Based EOT Detection

Encoder-based models excel in token-level sequence classification. WangchanBERTa (106M parameters) [26], pre-trained on large Thai corpora, captures Thai-specific linguistic phenomena like particles and implicit sentence boundaries. mDeBERTa-v3-base (276M parameters) [25] introduces disentangled attention mechanisms and multilingual coverage, supporting robust performance across Thai and English. These encoders are optimized for fine-tuning and efficient CPU inference, making them strong baselines for real-time EOT detection.

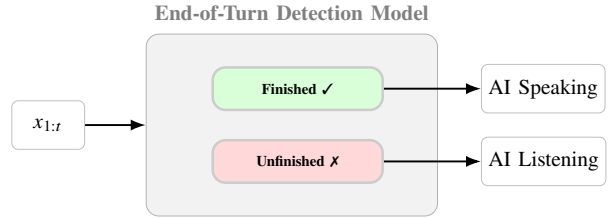

Fig. 2: Turn detection in full-duplex dialogue: given a streaming transcript $x_{1:t}$, the model decides whether the boundary at position $t$ is an *EOT (end)* or *Not-EOT (not-end)*, which then dictates if the AI starts speaking or continues listening.

An encoder-based paradigm approaches semantic end-of-turn detection as a single, holistic classification task applied to a complete unit of speech. [16], [17] The process generally unfolds as follows: a segment of dialogue—typically a full utterance captured after a pause—is prepared as the model's input sequence. [18] A special classification token is prepended to the sequence, which is then processed by an encoder model such as BERT [28]. The encoder operates bidirectionally, producing contextualized representations for all tokens, with the final hidden state of the [CLS] token serving as an aggregated semantic embedding of the entire input.

This representation is subsequently passed through a lightweight classification head, often implemented as a linear projection followed by a sigmoid or softmax activation. The output is a binary decision indicating whether the input segment constitutes an end-of-turn (EOT) or not-end-of-turn (non-EOT) [15]. By treating the problem in this way, encoder-based models rely on complete context windows, offering a global judgment on turn completion rather than the incremental, probabilistic assessments characteristic of decoder-based approaches.

## III. Task and Data

### A. Problem definition

Given a streaming transcript $x_{1:t}$, decide whether a boundary at position $t$ is an EOT (*end*) or *not-end* as shown in Figure 2. We treat subtitle line ends (and sentence-final punctuation when present) as positive labels.

### B. Dataset construction

We derive data from the YODAS corpus [6], filtering to retain only Thai utterances. Raw subtitles include substantial

noise: songs, advertisements, and non-dialogue content. We therefore applied several preprocessing steps:
- **Regex filtering:** Retain utterances containing Thai characters and within reasonable length bounds.
- **Noise filtering with LLM:** A large instruction-tuned model (Typhoon-v2.1-12B-Instruct [24]) was prompted to remove lines judged to be songs, advertisements, or otherwise irrelevant to conversational turn-taking.
- **Sentence segmentation:** The same LLM was further used to split subtitle lines into sentence-like units, improving alignment with potential turn boundaries.

*C. Labeling strategy*

For decoder models, each sentence is simply terminated with the model's existing EOT token, enabling direct likelihood estimation at run time. For encoder models, we convert each sentence into positive and negative examples: if the text length is sufficient, we cut it at the middle and label the first segment as *not-end* and the full sentence as *end*. This produces a balanced classification dataset without requiring explicit punctuation heuristics.

*D. Splits and statistics*

After preprocessing, the dataset contains approximately 59k Thai sentences. We split the data into 80% training, 10% validation and 10% testing.

## IV. MODELS

We evaluate four classes of models for text-only Thai End-of-Turn (EOT) detection as shown in Figure 3: (i) zero-shot and few-shot prompting with decoder-only models, which serve as a baseline to measure intrinsic model capabilities; (ii) zero-shot thresholding which determines the end of a conversational turn based on the probability of the native stop token, using a fixed threshold value; (iii) fine-tuned encoder-only models trained for explicit classification; and (iv) zero-shot thresholding with fine-tuned decoder-only models adapted through generative training. Across these paradigms, we compare the performance of Thai-specific "specialist" models against powerful multilingual "generalist" ones.

*A. Model Selection*

For the encoder paradigm, we selected two models to create a direct comparison.
- **WangchanBERTa-base** [26] is pre-trained exclusively on a massive Thai corpus. It was selected to test the hypothesis that specialized, in-domain knowledge is essential for capturing the nuanced linguistic features of EOT cues in Thai.
- **mDeBERTa-v3-base** [25] is our multilingual baseline. It allows us to determine if a more advanced generalist architecture can overcome a lack of language-specific pre-training to match or exceed the performance of the specialist model.

For the decoder paradigm, we selected models from two distinct scales—a compact and a mid-size version of each family—to assess how parameter count impacts performance and efficiency.
- **Llama3.2-Typhoon2-1B and Llama3.1-Typhoon2-8B** [24] are our Thai-specialist models. Based on modern Llama architectures, they have undergone extensive pre-training and instruction tuning on large corpora of Thai text, allowing us to assess the impact of model scale on this language-specific task.
- **Qwen3-0.6B and Qwen3-8B** [23] are chosen as their multilingual counterparts. These models are recognized for strong cross-lingual performance and architectural efficiency. Including both a compact (0.6B) and a mid-size (8B) version allows us to evaluate the trade-off between low-latency deployment and the higher accuracy offered by a larger parameter count.

*B. Decoder Zero-shot and Few-shot Prompting*

To evaluate zero-shot and few-shot performance of our selected decoder models, we frame Thai text-only EOT detection as an instruction-based binary classification task. In the zero-shot setting, the model is prompted to classify a text sequence as a completed or incomplete turn [20] (the full zero-shot template is shown in Figure 5). To assess the impact of in-context learning, the few-shot setting extends this with a small number of demonstrations [21]; the complete few-shot prompt is illustrated in Figure 6, covering common Thai conversational patterns.

*C. Encoder Model Training: Binary Classification*

The encoder models were fine-tuned on a binary classification task designed to explicitly identify sentence completion. We formulated the training data by labeling each complete instruction from the training set as "Finished". Concurrently, we generated one negative sample labeled "Unfinished" by truncating each instruction at its midpoint. This data augmentation strategy was designed to ensure the models learned the difference between partial and complete utterances. Both models were trained using the AdamW optimizer [29] with a learning rate of $2\times10^{-5}$. WangchanBERTa was trained for 5 epochs with a batch size of 256, while the larger mDeBERTa-v3-base was trained for 2 epochs with a batch size of 64. The best-performing checkpoint was selected based on the highest weighted F1-score on the held-out validation set.

*D. Decoder Model Training: Supervised Fine-Tuning*

In contrast, the decoder models were adapted using Supervised Fine-Tuning (SFT) with a causal language modeling objective. We fine-tuned the models exclusively on the complete, "Finished" utterances from the YODAS Thai dataset [6], using the field text and a maximum sequence length of 512 tokens. Training ran for a single epoch with per-device batch sizes of 16 for both training and evaluation (no gradient accumulation). We used the AdamW optimizer with 8-bit parameterization, a learning rate of $2 \times 10^{-5}$, weight decay 0.01, and a cosine learning rate scheduler with a 3% warmup; mixed precision was enabled with `bfloat16`. We evaluated and checkpointed every 100 steps. At inference time, turn completion was estimated by monitoring the probability assigned to the model's native stop token at each boundary and applying a threshold to this probability.

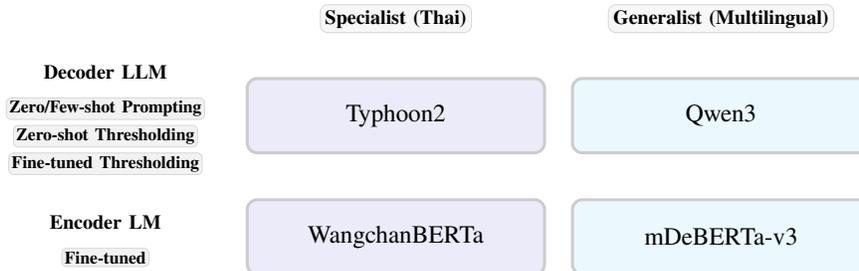

Fig. 3: Model taxonomy: zero-shot, few-shot prompting with decoder LLMs, zero-shot thresholding with decoder LLMs, fine-tuned decoder LLMs, and fine-tuned encoder LMs—crossed with specialist (Thai-specific) and generalist (multilingual) models.

*E. Decoder Zero-Shot Thresholding*

In addition to supervised fine-tuning, we also evaluated decoder models in a zero-shot setting. In this mode, the pre-trained model weights are left unchanged, and turn completion is estimated solely from the probability mass assigned to the model's native stop token at each boundary. A decision threshold is then applied to this probability. Rather than fixing the threshold arbitrarily, we select the operating point by analyzing the ROC curve on the validation set and choosing the threshold that maximizes Youden's $J$ statistic ($J = \text{TPR} + \text{TNR} - 1$) [22]. This approach provides a fair, model-agnostic comparison and ensures a balanced trade-off between true positive and false positive detections without requiring additional training.

## V. EXPERIMENTS

*A. Experimental Setup*

We evaluated all models on the test split of our preprocessed YODAS dataset [6], which was held out from training and validation. Our primary metric is the **F1-score** of the positive (*end*) class, which provides a balanced measure of precision and recall. We also report overall accuracy, precision, and recall for a comprehensive performance.

*B. Evaluation Methods*

We assessed model performance across three distinct paradigms:

- **Fine-tuning**: Models were explicitly trained on our Thai EOT dataset. Encoders were trained as binary classifiers, while decoders were trained with a causal language modeling objective to predict a stop token.
- **Instruction Prompting (Zero-shot and Few-Shot)**: Pre-trained decoder models were evaluated using an instruction prompt to classify turns without any weight updates. This was performed in both a **zero-shot** setting (no examples) and a **few-shot** setting (five examples). The full prompts are shown in Figures 5 and 6.
- **Zero-Shot Thresholding**: The raw probability of the pretrained decoder's native stop token was used as a score. An optimal decision threshold was then applied to classify the turn.

*C. Operating Points and Calibration*

For methods that output a probability score (fine-tuning and zero-shot thresholding), we report results at two operating points:

- **Uncalibrated (0.5)**: A fixed 0.5 threshold, showing out-of-the-box performance.
- **Calibrated (val-optimized)**: An optimal threshold $\tau^\star$ found by maximizing Youden's J-statistic on the validation set's ROC curve. This simulates a lightweight calibration step.

*D. Latency Measurement*

To assess real-time viability, average per-sample inference latency was measured on an **Intel® Xeon® Platinum 8480+** CPU with a batch size of one.

*E. Results and Analysis*

Our experimental results are summarized in Tables I, II, and III, with ROC curves presented in Figure 4. The findings reveal a clear performance hierarchy across the different methods and model types.

*a) Effectiveness of Supervised Fine-Tuning:* The primary finding is that supervised fine-tuning yields the highest performance. As shown in Table I, the fine-tuned **Llama3.2-Typhoon2-1B** [24] achieves the top F1-score of **0.881**, followed closely by the fine-tuned Qwen3-0.6B [23] (0.866) and the encoder-based mDeBERTa-v3-base [25] (0.861). These scores significantly outperform all zero-shot methods, demonstrating that while pretrained models have an intrinsic understanding of sentence structure, task-specific training is essential for reaching state-of-the-art performance in EOT detection.

TABLE I: Main results at validation-optimized thresholds (Youden's *J*) with CPU latency (batch=1) and parameter size. Latency measured on an Intel® Xeon® Platinum 8480+ CPU, averaged over 100 samples. Metrics are computed on the test set.

| Model | F1 | Acc | Prec | Rec | Time (s) | Size |
|---|---|---|---|---|---|---|
| **Zero-shot Thresholding** | | | | | | |
| Qwen3-0.6B | 0.622 | 0.570 | 0.571 | 0.684 | **0.09** | 0.6B |
| Qwen3-8B | 0.662 | 0.651 | 0.664 | 0.659 | 0.71 | 8B |
| Typhoon2-1B | 0.820 | 0.823 | **0.864** | 0.780 | 0.11 | 1B |
| Typhoon2-8B | 0.824 | 0.820 | 0.833 | 0.816 | 0.62 | 8B |
| **Fine-tuned** | | | | | | |
| Qwen3-0.6B | 0.866 | 0.861 | 0.861 | 0.872 | **0.09** | 0.6B |
| Typhoon2-1B | **0.881** | **0.874** | 0.860 | **0.902** | 0.11 | 1B |
| WangchanBERTa | 0.784 | 0.792 | 0.844 | 0.733 | 0.14 | 106M |
| mDeBERTa-v3-base | 0.861 | 0.855 | 0.853 | 0.870 | 0.29 | 276M |

TABLE II: Threshold sensitivity on the test set: *Calibrated (val-optimized via ROC/Youden's J)* vs. *Uncalibrated (fixed 0.5)*. $\tau^\star$ is the validation-optimized threshold.

| Model | $\tau^\star$ | Calibrated (val-optimized) | | | | Uncalibrated (0.5) | | | |
|---|---|---|---|---|---|---|---|---|---|
| | | F1 | Acc | Prec | Rec | F1 | Acc | Prec | Rec |
| **Zero-shot** | | | | | | | | | |
| Qwen3-0.6B | $3.47\times10^{-8}$ | 0.622 | 0.570 | 0.571 | 0.684 | 0.000 | 0.482 | 0.000 | 0.000 |
| Qwen3-8B | $2.60\times10^{-8}$ | 0.662 | 0.651 | 0.664 | 0.659 | 0.000 | 0.482 | 0.000 | 0.000 |
| Typhoon2-1B | $1.74\times10^{-6}$ | 0.820 | 0.823 | **0.864** | 0.780 | 0.000 | 0.482 | 0.000 | 0.000 |
| Typhoon2-8B | $2.58\times10^{-6}$ | 0.824 | 0.820 | 0.833 | 0.816 | 0.004 | 0.483 | **1.000** | 0.002 |
| **Fine-tuned** | | | | | | | | | |
| Qwen3-0.6B | 0.066 | 0.866 | 0.861 | 0.861 | 0.872 | 0.512 | 0.658 | 0.978 | 0.346 |
| Typhoon2-1B | 0.064 | **0.881** | **0.874** | 0.860 | **0.902** | 0.586 | 0.693 | 0.975 | 0.418 |
| WangchanBERTa | 0.570 | 0.784 | 0.792 | 0.844 | 0.733 | 0.791 | 0.791 | 0.821 | 0.763 |
| mDeBERTa-v3base | 0.521 | 0.861 | 0.855 | 0.853 | 0.870 | **0.861** | 0.854 | 0.848 | **0.875** |

TABLE III: Zero-/few-shot prompting results on the test set. Latency measured on an Intel® Xeon® Platinum 8480+ CPU, batch=1, averaged over 100 samples.

| Model | Shots | F1 | Acc | Prec | Rec | Time (s) |
|---|---|---|---|---|---|---|
| **Zero-shot Prompting** | | | | | | |
| Qwen3-0.6B | 0 | 0.316 | 0.483 | 0.671 | 0.483 | **0.179** |
| Qwen3-8B | 0 | **0.706** | **0.715** | 0.757 | **0.715** | 1.568 |
| Typhoon2-1B | 0 | 0.320 | 0.472 | 0.380 | 0.472 | 0.379 |
| Typhoon2-8B | 0 | 0.436 | 0.536 | 0.690 | 0.536 | 2.356 |
| **Few-shot Prompting** | | | | | | |
| Qwen3-0.6B | 5 | 0.362 | 0.500 | 0.638 | 0.500 | 0.259 |
| Qwen3-8B | 5 | 0.695 | 0.707 | **0.758** | 0.707 | 2.497 |
| Typhoon2-1B | 5 | 0.412 | 0.532 | 0.632 | 0.532 | 0.414 |
| Typhoon2-8B | 5 | 0.577 | 0.619 | 0.725 | 0.619 | 2.638 |

*b) Limitations of Zero-shot and Few-Shot Prompting:* Our evaluation of instruction-based prompting (Table III) shows this method is impractical for real-time EOT, regardless of the number of examples provided. The best F1-score achieved is 0.706 (Qwen3-8B, zero-shot). While adding five examples (few-shot) improves performance, the gains are inconsistent and failed to lift any model to a competitive level. More critically, the inference latency is prohibitively high (1.5–2.6 seconds) and **increases** with few-shot examples due to the longer input context. This makes instruction prompting, whether zero-shot or few-shot, orders of magnitude too slow for a real-time conversational agent.

*c) The Importance of Threshold Calibration:* The zero-shot thresholding method reveals a critical insight into decoder models. As seen in Table II, their out-of-the-box performance with a default 0.5 threshold is effectively zero (F1 ≈ 0.000). This is because a model's raw probability for a stop token is naturally very low. However, after **calibrating the threshold** on a validation set, performance becomes respectable, with Llama3.1-Typhoon2-8B reaching an F1-score of 0.824. This highlights that while zero-shot thresholding is feasible, it is not a "plug-and-play" solution and mandates a data-driven calibration step. In contrast, the fine-tuned encoder models are naturally well-calibrated for classification and perform robustly without this step.

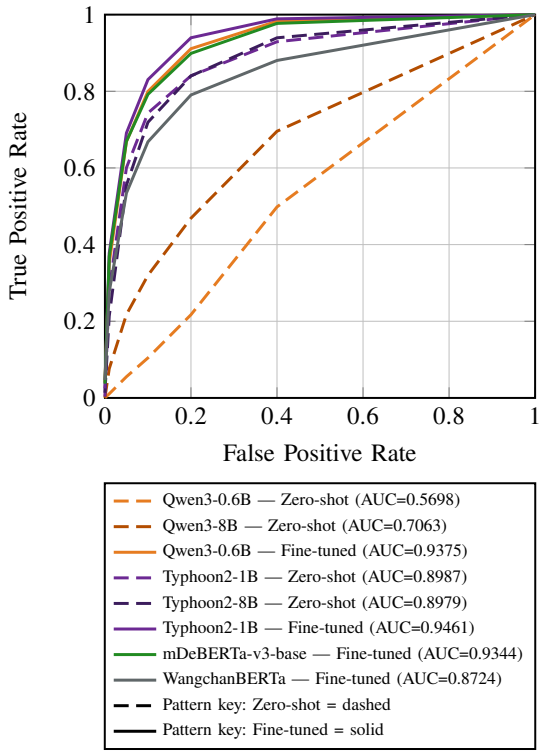

Fig. 4: ROC curves for Thai EOT detection models with thresholding on the test set. Color encodes model family and size (darker = larger model); line pattern encodes training regime (dashed = zero-shot, solid = fine-tuned).

*d) Comparing Specialist and Generalist Models:* The comparison between specialist (Thai) and generalist (multilingual) models produced nuanced results. In zero-shot scenarios, the Thai-specialist Typhoon models clearly outperform the generalist Qwen3 models, suggesting that language-specific pretraining is key for understanding turn-taking cues without fine-tuning. However, after fine-tuning, the powerful architecture of the generalist **mDeBERTa-v3-base (0.861 F1) surpassed the Thai-specialist WangchanBERTa [26] (0.784 F1)**. This indicates that for this classification task, a more modern and capable model architecture can be more important than language-specific pretraining, provided sufficient fine-tuning data is available.

## VI. Discussion

Our experimental analysis provides clear, actionable insights for developing a real-time Thai EOT detection system. The results point to an optimal balance of accuracy and latency.

*a) Identifying the Optimal Accuracy-Latency Trade-off:* For production systems, the central challenge is balancing performance with efficiency. Our results (Table I and Figure 1) show that the fine-tuned **Llama3.2-Typhoon2-1B** [24] strikes an optimal balance. It delivers the highest accuracy (0.881 F1) while maintaining a low CPU latency of 110ms, making it well-suited for real-time voice agents. For applications on highly constrained devices, the fine-tuned **Qwen3-0.6B** [23] is an excellent alternative, offering nearly comparable accuracy (0.866 F1) at an even faster 90ms.

*b) Deployment Considerations: Encoders vs. Decoders:* The choice between an encoder and a decoder model involves a trade-off between deployment simplicity and maximum performance.

- **For Peak Performance**: A fine-tuned decoder like Typhoon2-1B is preferable. It achieves the highest F1-score but requires a calibration step on a validation set to determine the optimal decision threshold.
- **For Simplicity and Robustness**: A fine-tuned encoder like **mDeBERTa-v3-base** [25] offers a compelling alternative. It provides strong, reliable performance that is not sensitive to the decision threshold, making it a "drop-in" solution that works well out-of-the-box.

*c) The Role of Thai-Specific Pretraining:* Our findings indicate that Thai-specific pretraining is most critical when task-specific fine-tuning is not possible. In such zero-shot scenarios, a model like Typhoon2 is the only viable option. However, if a dataset for fine-tuning is available, the underlying power of the model architecture becomes a more dominant factor, allowing a strong generalist model to excel.

*d) Advancing Beyond Silence-Based Endpointing:* This work demonstrates that a lightweight, fine-tuned transformer can provide a fast and accurate semantic EOT signal. This approach avoids the arbitrary delays of silence-based endpointing and correctly interprets linguistically complete utterances even when they contain pauses. By integrating such a model, conversational AI systems can reduce response latency and achieve more natural, fluid turn-taking, significantly enhancing the user experience.

## VII. Limitations and Ethics

Our labels inherit subtitle biases and timing drift; true conversational EOT may diverge from subtitle line breaks. We do not model acoustic cues (prosody, overlap), which matter in multi-party settings. Banking dialogs may include sensitive content; dataset handling must comply with privacy policies.

## VIII. Conclusion

We introduce a Thai-focused, text-only EOT formulation and compare fine-tuned small transformers against zero-/few-shot LLMs for real-time agents. This establishes a practical baseline and complements audio-native turn detectors in open-source stacks. Future work will integrate lightweight prosodic features and extend to multi-party overlap.

ACKNOWLEDGMENTS

The authors would like to express their sincere gratitude to Weerin Chantaroje and Tutanon Sinthuprasith for their guidance and continuous support in providing the necessary resources for this work. We also wish to thank our colleagues, Peerawat Rojratchadakorn and Nut Chukamphaeng, for their helpful feedback and for sharing their experience relevant to this project. Additional thanks go to our colleagues in the SCBX R&D and Innovation Lab teams for their feedback, particularly on Thai conversational examples. Computing resources were generously provided by Hatari, which greatly facilitated our experiments and analyses.

APPENDIX

In this appendix, we provide the full text of the prompts used in our experiments. We include both zero-shot and few-shot variants.

*a) Zero-Shot Prompt.:* The zero-shot prompt (Figure 5) provides only task instructions and output constraints. It emphasizes clarity and single-token outputs, serving as a baseline for detecting complete vs. incomplete turns.

---

**Zero-Shot Prompting**

```
You are an expert in Thai conversational
analysis. A user's turn is considered
'Complete' if they finish a sentence,
ask a question (e.g., ใช่ไหม), or use a
concluding particle like ครับ, ค่ะ, or นะ. A
turn is 'Incomplete' if it ends abruptly
mid-thought or trails off. Analyze the
transcript and respond with Complete or
Incomplete. Your answer must be a single
token.
Text: { {TRANSCRIPT_PREFIX} } Label:
```

---

Fig. 5: The full zero-shot prompt

*b) Few-Shot Prompt.:* The few-shot prompt (Figure 6) extends the zero-shot version with positive and negative examples. These cover common conversational patterns and help reduce ambiguity in borderline cases.

---

**Few-Shot Prompting**

```
You are an expert in Thai conversational
analysis. A user's turn is considered
'Complete' if they finish a sentence,
ask a question (e.g., ใช่ไหม), or use a
concluding particle like ครับ, ค่ะ, or นะ. A
turn is 'Incomplete' if it ends abruptly
mid-thought or trails off. Analyze the
transcript and respond with Complete or
Incomplete. Your answer must be a single
token.
Examples:
Text: ช่วยเปิดบัญชีให้หน่อยครับ Label: Complete
Text: วันนี้โอนเงินได้ถึงกี่โมงครับ Label: Complete
Text: ขอโอนเงินไปที่บัญชีคุณแม่แล้วก็ Label: Incomplete
Text: เดี๋ยวขอเช็คยอดก่อนนะ แล้ว Label: Incomplete
Text: งั้นตัดบัตรเครดิตใบนี้เลยครับ Label: Complete
Text: { {TRANSCRIPT_PREFIX} } Label:
```

---

Fig. 6: The full few-shot prompt